\documentclass{article}

\usepackage[main, final]{style}

\usepackage[utf8]{inputenc} 
\usepackage[T1]{fontenc}    
\usepackage{hyperref}       
\usepackage{url}            
\usepackage{booktabs}       
\usepackage{amsfonts}       
\usepackage{nicefrac}       
\usepackage{microtype}      
\usepackage{xcolor}         
\usepackage{xspace}

\usepackage{amsmath}
\usepackage{graphicx}

\newcommand{\papernamelong}{Coarse-to-Real}
\newcommand{\papernameabbr}{C2R\xspace}

\title{\papernamelong: Generative Rendering for Populated Dynamic Scenes}

\author{Gonzalo Gomez-Nogales \\
Universidad Rey Juan Carlos \\
Móstoles, Spain \\
\and
Yicong Hong \\
\and 
Chongjian Ge \\
Adobe Research \\
San Jose, USA \\
\and 
Peiye Zhuang \\
Roblox \\
San Mateo, USA \\
\and 
Marc Comino-Trinidad \\
Universidad Rey Juan Carlos \\
Móstoles, Spain \\
\and 
Dan Casas \\
Universidad Rey Juan Carlos \\
Móstoles, Spain \\
\and 
Yi Zhou \\
Roblox \\
San Mateo, USA \\
}

\begin{document}

\maketitle

\begin{figure}[h]
\vspace{-1.2cm}
\includegraphics[width=\linewidth]{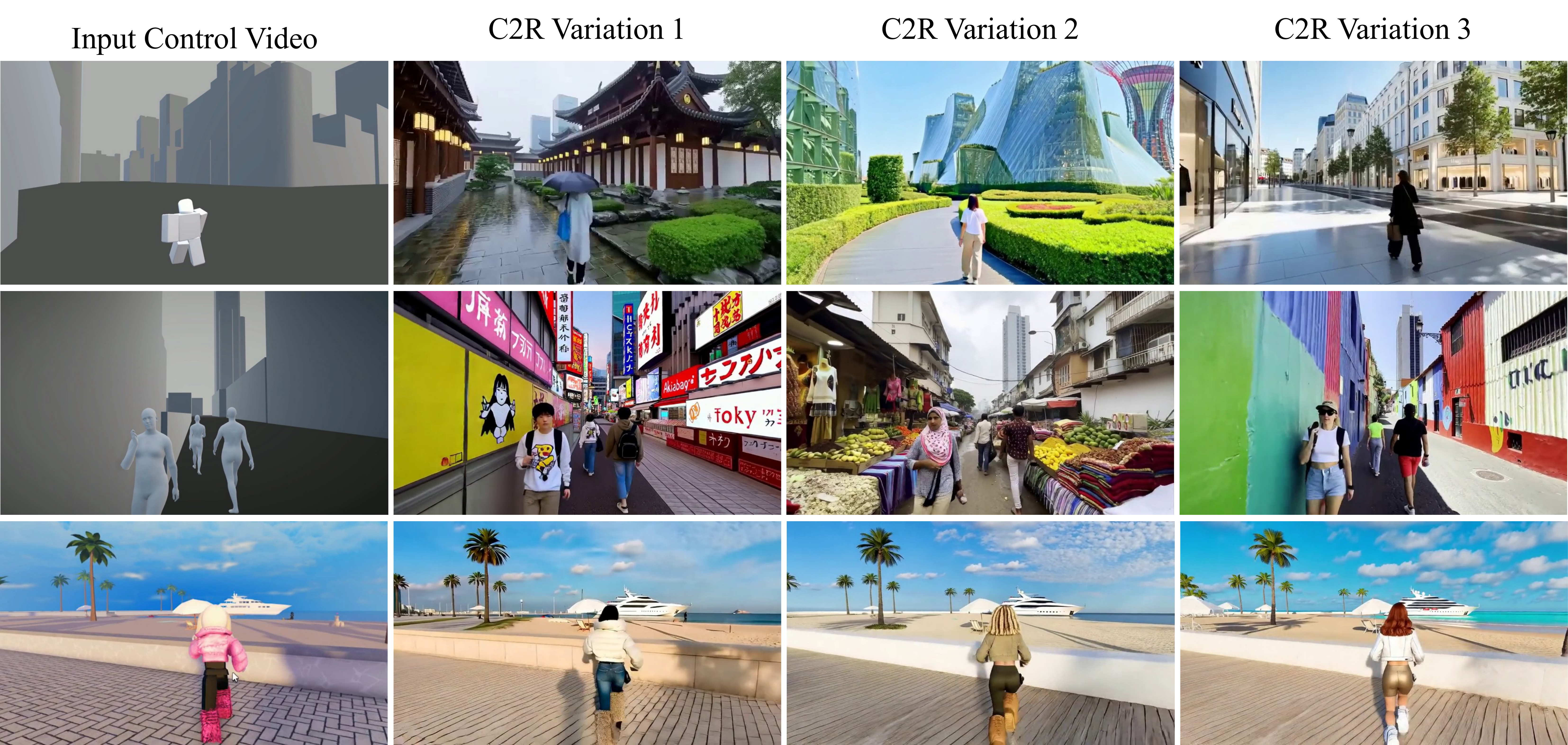}
\caption{\textbf{\papernameabbr (\papernamelong):} transforms a coarse 3D control video from different types (left col.) like coarse block-like low-poly characters (top), humanoid coarse models (middle) and game videos (bottom), into diverse scenes via text prompts, varying texture detail, lighting, weather, location, dynamics and clothing styles while maintaining consistent camera motion and human trajectories.}
\vspace{0.2cm}
\label{fig:teaser}
\end{figure}

\begin{abstract}
\label{sec:abstract}
Traditional rendering pipelines rely on complex assets, accurate materials and lighting, and substantial computational resources to produce realistic imagery, yet they still face challenges in scalability and realism for populated dynamic scenes. 
We present \papernameabbr (\papernamelong), a generative rendering framework that synthesizes real-style urban crowd videos from coarse 3D simulations. Our approach uses coarse 3D renderings to explicitly control scene layout, camera motion, and human trajectories, while a learned neural renderer generates realistic appearance, lighting, and fine-scale dynamics guided by text prompts. To overcome the lack of paired training data between coarse simulations and real videos, we adopt a two-stage synthetic-real domain-hedging strategy that first learns a strong generative prior from large-scale real footage, then introduces controllability by using a small amount of paired synthetic coarse--fine data to anchor shared implicit spatio-temporal features across domains. The resulting system supports coarse-to-fine control, generalizes across diverse CG and game inputs, and produces temporally consistent, controllable, and realistic urban scene videos from minimal 3D input.
We will release the model and project webpage.
\end{abstract}

\vspace{-0.3cm}
\section{Introduction}
\label{sec:introduction}
\vspace{-0.2cm}

Rendering populated realistic urban scenes with dynamic camera trajectories and complex group motion remains expensive and difficult in traditional computer graphics (CG) pipelines.
Achieving realism typically requires high-quality assets, detailed materials, accurate lighting, and carefully tuned simulations, leading to substantial production cost and memory overhead, yet often still exhibiting a gap from real-world appearance.
Recent advances in generative text- and image-to-video models offer an alternative \cite{ho2022video,chen2023videocrafter1,yang2025cogvideox,ruan2022mmdiffusion,singer2023makeavideo}, but existing methods struggle with scenes involving multiple interacting humans, long-term temporal consistency, and coherent camera motion.
Purely text-based control is insufficient for precisely specifying scene structure and dynamics, and image-conditioned approaches \cite{geng2025motionprompting}, like Sora, remain limited to simple and similar camera positions for scenes involving complex crowd dynamics, as shown in the Appendix Section~\ref{app:existing_models}.
Video-to-video models \cite{vid2vid-zero,nvidia2025worldsimulationvideofoundation,cheng2024consistent,li2024vidtome} combined with ControlNet \cite{zhang2023adding} offer promising and more controllable alternatives, but they usually use strong geometric control signals that over-constrain the generation and the expressivity of the output video.

A key observation motivating this work is that many challenges related to dynamic video generation are trivial in 3D, while many of the most expensive aspects of 3D rendering are naturally handled by data-driven models. 
Camera motion, spatial structure, and temporal consistency are explicit and stable in coarse 3D scenarios, whereas detailed geometry, textures, materials, lighting, and visual diversity are costly to author and render.
This complementary relationship motivates a hybrid paradigm, where coarse 3D simulation provides structural and dynamic control, and a learned generative model functions as a renderer that synthesizes realistic appearance.

We introduce \textbf{\papernameabbr (\papernamelong)}, a generative rendering framework that produces real-style populated urban videos from coarse 3D inputs. \papernameabbr takes temporally consistent coarse renderings encoding scene layout, camera trajectories, and human motion, and enriches them with realistic textures, lighting, and fine-scale dynamics learned from real-world video data. The framework offers flexible control, including inpainting clothing, hair, buildings, and environmental details beyond the input structure, adjusting the strength of generative rendering, and supporting coarse-to-fine inputs. Importantly, \papernameabbr is agnostic to specific human templates and asset formats, allowing it to generalize across a wide range of CG, game, and animation scenes.

A central technical challenge is the absence of paired training data between coarse CG simulations and real-world videos. To address this challenge, we propose a synthetic-real domain-hedging strategy based on a two-stage learning framework. In the first stage, the model is trained on large-scale unpaired real-world videos to learn a strong photorealistic generative prior. In the second stage, controllability is introduced by grounding this prior using \emph{implicit spatio-temporal features} extracted from both real and synthetic inputs, enabling the model to interpret structural cues in a shared feature space across domains despite the absence of explicit pairing, and to adaptively hallucinate content based on the level of detail present in the input. A small proportion of synthetic paired coarse–fine data anchors the correspondence between coarse structure and realistic appearance, while the dominance of real data prevents contamination by CG-specific artifacts.
To support diverse generation conditions, we curate and annotate footage from five continents, covering a wide range of cities, weather conditions, lighting, and clothing styles. We evaluate different latent feature insertion strategies and demonstrate that \papernameabbr produces temporally consistent, controllable, and realistic urban scene renderings from minimal 3D simulation.

\vspace{-0.3cm}
\section{Related Works}
\vspace{-0.2cm}

We review research relevant to the high-quality synthesis of populated urban videos, from the perspective of traditional computer graphics pipelines (Sec.\ref{sec:work-cg}), video and world models (Sec.\ref{sec:work-video}) and controllable video generation techniques (Sec.\ref{sec:work-condition}). Our discussion emphasizes the unique challenges of maintaining structural control and temporal consistency in dynamic scenes with complex crowd interactions. 

\vspace{-0.3cm}
\subsection{Traditional CG for Dynamic Populated Scenes} 
\label{sec:work-cg}

Rendering populated urban environments with dynamic camera trajectories and complex group motion is among the most resource-intensive tasks in traditional computer graphics pipelines. Achieving high-fidelity realism necessitates extensive libraries of quality assets, including diverse 3D human meshes, detailed architecture, and physically based materials. Generating these components typically requires expert manual modeling or costly 3D scanning~\cite{loper2023smpl,anguelov2005scape}. As scene complexity scales, the memory overhead for high-resolution textures and geometry strains computational resources and compromises rendering efficiency. Beyond static assets, animating large crowds requires simulating intricate group behaviors and social interactions~\cite{helbing1995social}. While traditional techniques like skeletal animation and motion graphs~\cite{kovar2023motion} provide stability, they often struggle to produce naturalistic motion and vivid dynamics. Consequently, significant visual discrepancies compared to real-world appearances persist—particularly for large-scale crowds with diverse clothing and hair patterns—rendering the scalable authoring of realistic urban scenes practically challenging. Unlike these resource-intensive pipelines, our \papernameabbr framework leverages coarse 3D simulations as lightweight structural proxies, delegating the synthesis of photorealistic details and vivid dynamics to a data-driven generative renderer.

\vspace{-0.3cm}
\subsection{Video and World Models} 
\label{sec:work-video}

Recent advances in text-to-video \cite{ho2022video,chen2023videocrafter1,yang2025cogvideox,ruan2022mmdiffusion,wan2025wanopenadvancedlargescale,singer2023makeavideo} and image-to-video \cite{esser2023structure,chen2023mcdiff} generation offer a compelling alternative by synthesizing realistic imagery directly from data.
However, existing generative video models struggle to reliably produce scenes with multiple interacting humans, long-term temporal consistency, or coherent camera motion. In Figure~\ref{fig:defaults} in the Appendix Section~\ref{app:existing_models}, SORA and WAN 2.1~\cite{wan2025wanopenadvancedlargescale} show limited controllability over human motion and camera trajectories, and tend to generate similar viewing angles across populated urban scenes.
Moreover, purely text-based control is insufficient for precisely specifying scene structure, camera trajectories, and crowd dynamics, while image-conditioned approaches inherit the limitations of the input frame and offer limited control over motion and layout. 
As a result, current video generation systems fall short as practical tools for authoring structured, dynamic urban scenes.

To provide the structural grounding required for such urban scenes, a parallel line of research has emerged around world models. Unlike purely pixel-based generators, these models aim to leverage explicit 3D priors~\cite{kerbl20233d,wang2024dust3r} to build consistent traversable scenes. For instance, WorldGen~\cite{wang2025worldgentexttraversableinteractive} utilizes procedural generation to ensure global layout stability and precise camera control. However, while these models excel at maintaining static structure, they often struggle to capture the vivid, stochastic dynamics of populated urban crowds. 

Unlike these approaches, \papernameabbr\ does not attempt to reconstruct a 3D world or rely on purely procedural logic. Instead, we use coarse 3D simulations as structural proxies and delegate the computationally expensive rendering of photorealistic textures, lighting, and fine-grained crowd dynamics to a learned generative model. This hybrid formulation allows \papernameabbr\ to bypass the "sim-to-real" gap that persists in traditional CG and the lack of structural control in monolithic video models.

\vspace{-0.3cm}
\subsection{Controllable Video Generation} \label{sec:work-condition}

To bridge the gap between generative realism and structural precision, several lines of work have explored explicit conditioning for video synthesis: (1) One of the strategies focuses on sparse motion and camera guidance. Methods such as CameraCtrl \cite{he2025cameractrl}, GEN3C \cite{ren2025gen3c}, and Wan-Move \cite{chu2025wanmove} incorporate explicit camera parameters or trajectories to guide the denoising process. Similarly, motion-driven approaches like Motion Prompting \cite{geng2025motionprompting} and MCDiff \cite{chen2023mcdiff} utilize user-defined strokes or points to specify movement. While effective for single-subject scenes, these methods often struggle with the combinatorial complexity of crowded environments, where multiple  characters interact with independent and varying velocities. Furthermore, sparse signals often fail to ground the global scene layout as robustly as a 3D proxy. (2) Another strategy involves dense structural conditioning, often referred to video-to-video (vid2vid) synthesis. Existing frameworks \cite{esser2023structure,wang2018vid2vid,nvidia2025worldsimulationvideofoundation} use per-frame geometric cues, e.g., depth maps, Canny edges, or semantic layouts, to transform input sequences. However, these strong geometric signals often over-constrain the model expressivity, forcing the output to strictly follow the input geometry and preventing the generative renderer from adding "vivid dynamics".

Our \papernameabbr\ takes a different path by leveraging an implicit spatial-temporal feature representation. Unlike previous methods that rely on either sparse trajectories or rigid dense maps, \papernameabbr\ utilizes coarse 3D representations as lightweight structural proxies. This allows our framework to maintain the global stability and intent of the 3D scene while giving the generative model the freedom to generate high-fidelity textures, physically plausible lighting, and fine-grained crowd dynamics learned from real-world footage.

\begin{figure}[t]
  \centering
  \includegraphics[width=\linewidth]{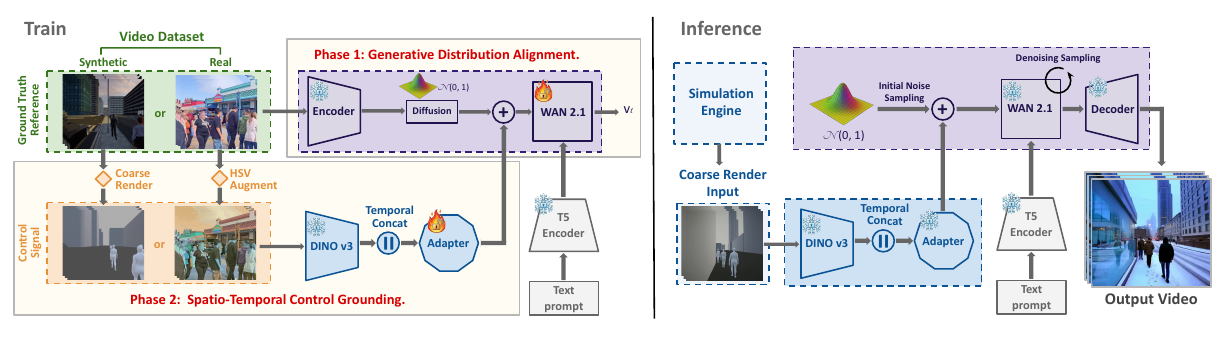}
  \caption{\textbf{Overview of \papernameabbr}
  A two-stage framework decouples photorealistic prior learning from structural control: the diffusion backbone is first adapted to real videos, then grounded through a synthetic-real domain-hedging stage that uses implicit spatio-temporal features shared across real and synthetic data. At inference, a coarse render and text prompt guide denoising to synthesize realistic videos following input motion and layout.}
  \label{fig:pipeline}
\vspace{-0.3cm}
\end{figure}

\section{Method}
\label{sec:method}

\vspace{-0.3cm}
\subsection{Overview and Design Choices}
\label{subsec:overview_design}

Our goal is to generate vivid and real-styled videos that follow the camera motion and scene dynamics of coarse 3D or game-engine renders. This setting introduces three challenges: (1) coarse renders provide reliable structure but lack realistic appearance; (2) paired supervision between coarse 3D inputs and photorealistic targets is only available for CG data, while real videos are abundant but lack aligned control signals, requiring a bridge between real and synthetic inputs; and (3) explicit structural controls (e.g., depth, edges, poses) over-constrain generation by entangling appearance with structure, often leading to \emph{feature leakage}, where the model bypasses synthesis and reconstructs the control input.

To address these challenges, we decouple photorealistic prior learning from structural controllability using a two-stage training strategy. We first learn a high-fidelity generative prior from large-scale real videos without structural conditioning. We then introduce controllability through a domain-hedging stage, where \emph{implicit spatio-temporal features} extracted from both real and synthetic inputs are mapped into a shared feature space via a lightweight adapter. This design yields adaptive behavior: coarse inputs allow flexible synthesis, while richer inputs naturally enforce stronger adherence to motion and layout.

\vspace{-0.3cm}
\subsection{Architecture Preliminaries}
\label{subsec:preliminaries}

We build on WAN 2.1~\cite{wan2025wanopenadvancedlargescale} and denote its VAE encoder and decoder by $E$ and $D$, and its diffusion backbone (Diffusion Transformer) by $\mathrm{DiT}$ with parameters $\theta$. Text conditioning is provided by a pretrained T5-XXL encoder $\mathrm{T5}$  \cite{raffel2020exploring} that maps a prompt $\mathbf{c}$ to an embedding $\mathbf{e}_{\text{text}}=\mathrm{T5}(\mathbf{c})$. For control signals, we extract patchwise features using a pretrained DINOv3 ViT~\cite{simeoni2025dinov3}, and train an adapter $A$ that projects these features into the diffusion latent space.

\vspace{-0.3cm}
\subsection{Stage I: Generative Distribution Alignment}
\label{subsec:stage1_gda}

\textbf{Motivation.} Coarse control signals alone cannot teach photorealistic appearance. We therefore first adapt the diffusion backbone to the target ``realistic urban video'' distribution so that it learns lighting, materials, and natural motion before being asked to follow coarse structure.

\textbf{Training.} In this stage, we fine-tune only the diffusion backbone $\theta$ on real-world videos, while freezing the VAE $E,D$ and the text encoder. 
Given a real video $\mathbf{x}$ and its corresponding prompt embedding $\mathbf{e}_{\text{text}}$, we first encode the video to latent representation:
\begin{equation}
\mathbf{z}_{0} = E(\mathbf{x}).
\end{equation}
Then, following the Flow Matching formulation applied in Wan 2.1, we sample a timestep $t \in [0, T]$ and apply the standard diffusion process by adding Gaussian noise ${\epsilon} \sim \mathcal{N}({0}, {I})$ to obtain the noisy latents $\mathbf{z}_{t}=(1-t)\mathbf{z}_{0}+t{\epsilon}$. The model is trained to predict the velocity field:
\begin{equation}
\mathbf{v}_{t} = \mathrm{DiT}_{\theta}(\mathbf{z}_{t}, t, \mathbf{e}_{\text{text}}),
\end{equation}
where the objective is defined as:
\begin{equation}
\boldsymbol{L}_{\text{FM}} =
\mathbb{E}_{\mathbf{z}_0, t, \mathbf{e}_\text{text}}
\Bigl[
\bigl\|
\text{DiT}_\theta(\mathbf{z}_t, t, \mathbf{e}_\text{text}) - (\epsilon-\mathbf{z}_{0})
\bigr\|_2^2
\Bigr].
\label{eq:diff_objective}
\end{equation}
At inference, we can predict the denoised latents $\hat{\mathbf{z}}_{0}$ through an iterative sampling process, and the final clean video results can be obtained by the vae decoder via $\hat{\mathbf{x}} = D(\hat{\mathbf{z}}_{0})$.

\vspace{-0.3cm}
\subsection{Stage II: Spatio-Temporal Control Grounding}
\label{subsec:stage2_ctrl}

\textbf{Motivation.} After Stage I, the model generates realistic videos but lacks controllability. Stage II teaches the model to interpret coarse structure \emph{without} forcing an explicit signal (e.g., depth/edges) as in ControlNet \cite{zhang2023adding} that would rigidly constrain silhouettes. Instead, we use \emph{implicit} features from a self-supervised encoder that preserve layout and motion while being less tied to pixel appearance.

\textbf{Training.} We freeze the tuned diffusion backbone 
and train only the control pathway, i.e., the adapter $A$ (and any lightweight temporal aggregation described below). Let $\mathbf{x}_{\text{ctrl}}$ be the control-branch input. We extract features frame-by-frame:
\begin{equation}
\mathbf{f}_i = \mathrm{DINO}(\mathbf{x}_{\text{ctrl}}^{(i)}), \quad i=1,\dots,N,
\end{equation}
and construct a spatio-temporal representation by temporal concatenation:
\begin{equation}
\mathbf{f}_{1:N} = \mathrm{Concat}_i(\mathbf{f}_1,\dots,\mathbf{f}_N).
\end{equation}
The adapter projects these features to a latent guidance tensor:
\begin{equation}
\hat{\mathbf{z}}_{\text{ctrl}} = A(\mathbf{f}_{1:N}).
\end{equation}
We inject the control guidance into the diffusion latent by element-wise addition:
\begin{equation}
\hat{\mathbf{z}}_{\text{t}} = \mathbf{z}_{\text{t}} + \hat{\mathbf{z}}_{\text{ctrl}}.
\label{eq:additive_fusion}
\end{equation}
Our experiments show that simple feature addition matches or outperforms multi-head cross-attention for integrating control features. 
Similar to Stage I, the diffusion backbone then denoises $\hat{\mathbf{z}}_{\text{t}}$ conditioned on text, and we optimize $A$ with the same diffusion objective (Eq.~\ref{eq:diff_objective}) on the corresponding target video.

\vspace{-0.3cm}
\subsection{Synthetic-Real Domain Hedging}
\label{subsec:domain_hedging}

A controllable video generation model would ideally be trained on paired data, where a coarse 3D input video is matched with a photorealistic target. In practice, such paired supervision can only be obtained from computer-generated (CG) data, which is expensive to produce, limited in diversity, and constrained by rendering fidelity. In contrast, real-world videos are abundant and available at a vastly larger scale, exhibiting rich variations in appearance, motion, and camera trajectories that are difficult to reproduce synthetically. To exploit this diversity and improve photorealism, we adopt a synthetic-real domain-hedging strategy, training \papernameabbr with \emph{orders of magnitude more real videos than synthetic ones}, using real data to shape a strong photorealistic generative prior and synthetic data to provide sparse but essential paired supervision.

To bridge the domain gap between unpaired real and paired synthetic data, we incorporate a step within our training pipeline that maps both real videos and synthetic coarse renders into a \emph{common spatio-temporal feature space}. In this space, control signals extracted from real videos and from coarse 3D simulations are encouraged to be compatible, enabling joint training despite their very different origins. Paired synthetic data anchors this space by linking coarse geometry to high-quality target appearance, while real videos densely populate it and enrich the learned generative distribution. Please refer to the Appendix Section~\ref{app:data} for details on the collection and annotation of the synthetic and real datasets.

\vspace{-0.3cm}
\subsection{Adaptive Spatio-Temporal Control from Implicit Features}
\label{subsec:implicit_control}

A central challenge in controllable video generation is extracting spatio-temporal control signals that generalize across real and synthetic domains. Explicit structural representations such as depth, edges, optical flow, or poses~\cite{vid2vid-zero,nvidia2025worldsimulationvideofoundation,cheng2024consistent,li2024vidtome} often entangle geometry with appearance, especially in real videos rich in texture and fine detail. As a result, they tend to over-constrain generation and encourage direct reconstruction rather than synthesis.

To address this, we adopt an \emph{implicit} spatio-temporal representation that preserves scene layout and motion while abstracting away low-level appearance. We extract dense patch-level features using a pretrained self-supervised vision encoder~\cite{simeoni2025dinov3} and aggregate them temporally to form a spatio-temporal feature grid. These features (visualized in Figure~\ref{fig:dino_viz}) are appearance-robust and semantically structured, making them suitable as a shared control representation for both real videos and synthetic coarse renders.

To further reduce appearance leakage when using real videos, we apply a video-consistent HSV transformation to the control branch during training. This suppresses color and texture correlations while preserving geometric structure and temporal coherence, encouraging the control signal to focus on layout and dynamics rather than pixel fidelity. Details can be found in the Appendix Section~\ref{paragraph:hsv_decorrelation}.

This implicit formulation naturally yields \emph{adaptive control}. When the input is very coarse, the extracted features provide high-level structural cues, allowing the model to freely hallucinate fine-scale details and secondary motion. When richer structural information is present, the same mechanism enforces stronger adherence to the input motion and spatial configuration. As a result, \papernameabbr supports a wide range of inputs—from low-poly game-engine videos to more detailed simulations—without requiring different control strategies or hand-crafted rules.


\vspace{-0.3cm}
\subsection{Inference}
\label{subsec:inference}


As shown in Fig.~\ref{fig:pipeline}, at inference time, given a coarse control video $\mathbf{x}_{\text{coarse}}$ and a text prompt $\mathbf{c}$, we compute a control latent $\hat{\mathbf{z}}_{\text{ctrl}} = A(\mathrm{DINO}(\mathbf{x}_{\text{coarse}}))$ and a text embedding $\mathbf{e}_{\text{text}} = \mathrm{T5}(\mathbf{c})$. We then perform standard diffusion sampling using a Flow Matching (Rectified Flow) sampler, which is solved with a first-order Euler integrator, starting from Gaussian noise $\mathbf{z}_{T} \sim \mathcal{N}(0, I)$. At each denoising step, control is injected via additive fusion (Eq.~\ref{eq:additive_fusion}), while textual conditioning is incorporated through cross-attention. After the final step, the decoded output is obtained as $\hat{\mathbf{x}} = D(\hat{\mathbf{z}}_{0})$.

\vspace{-0.3cm}
\subsubsection{Guidance control}
To balance structural faithfulness and textual alignment, we use Adaptive Prompt Guidance (APG)~\cite{castillo2025ag} during sampling. APG dynamically scales conditioning contributions, allowing the model to deviate from coarse input appearance (e.g., colors/textures) while maintaining motion and layout.
\section{Experiments}


In this section, we present a comprehensive evaluation of C2R to assess its performance in synthesizing high-fidelity, controllable urban videos. We utilize a training dataset consisting of 240k clips of real-world footage and 1.3k clips of synthetic data, with specific details regarding collection and annotation provided in the Appendix Section~\ref{app:data}. Our evaluation begins by introducing the quantitative metrics used to measure both semantic alignment and structural fidelity. To further validate our architectural and data-driven design choices, we conduct ablations on the real-to-synthetic sampling ratio in our domain-hedging strategy, our additive feature injection strategy, and the necessity of HSV decorrelation. We then provide qualitative evidence of the model's adaptive capacity to handle varied levels of input geometry, ranging from low-poly simulations to more detailed renders. Finally, we compare our framework against five publicly available state-of-the-art baselines to demonstrate our improvements in visual expressivity.

\subsection{Metrics}

We evaluate our method along two complementary axes: \emph{text–video alignment} and \emph{structural consistency}. To measure how well the generated video follows the input motion and structure, we use VE-Bench~\cite{sun2025ve}, which is designed for video editing and control-based evaluation. While VE-Bench is applicable to our setting, it may penalize large appearance changes and favor conservative edits that closely preserve the source. To account for this limitation, we additionally report VQAScore~\cite{lin2024evaluating}, a text–video alignment metric that directly evaluates prompt adherence. Using both metrics together provides a more balanced quantitative assessment of controllability and semantic alignment.

\subsection{Ablations}



\begin{figure}[t]
  \centering

  \begin{minipage}[t]{0.40\linewidth}
    \vspace{0pt}
    \centering
    \textbf{(a) Quantitative results}
    \vspace{0.35em}

    {\scriptsize
    \setlength{\tabcolsep}{2.0pt}
    \renewcommand{\arraystretch}{0.88}
    \resizebox{0.98\linewidth}{!}{%
    \begin{tabular}{lrr}
      \toprule
      Model & VE-Bench$\uparrow$ & VQAScore$\uparrow$ \\
      \midrule
      WAN2.1 + CN & 0.0086 & 0.0478 \\
      \midrule
      100\% synth. & 0.4675 & \textbf{0.5242} \\
      50\% real + 50\% synth. & 0.4370 & \underline{0.5120} \\
      70\% real + 30\% synth. & 0.4560 & 0.4538 \\
      90\% real + 10\% synth. & 0.5020 & 0.4438 \\
      95\% real + 5\% synth. & 0.4809 & 0.4262 \\
      99\% real + 1\% synth. & \underline{0.5317} & 0.3420 \\
      100\% real & \textbf{0.8090} & 0.1743 \\
      \midrule
      0 head & 0.5317 & \underline{0.3420} \\
      1 head & \underline{0.5837} & \textbf{0.3536} \\
      10 heads & \textbf{0.7893} & 0.0046 \\
      \bottomrule
    \end{tabular}%
    }
    }
  \end{minipage}
  \hfill
  \begin{minipage}[t]{0.57\linewidth}
    \vspace{0pt}
    \centering
    \textbf{(b) Ablation trends}
    \vspace{0.35em}

    \includegraphics[
      width=\linewidth,
      trim=5 5 5 5,
      clip
    ]{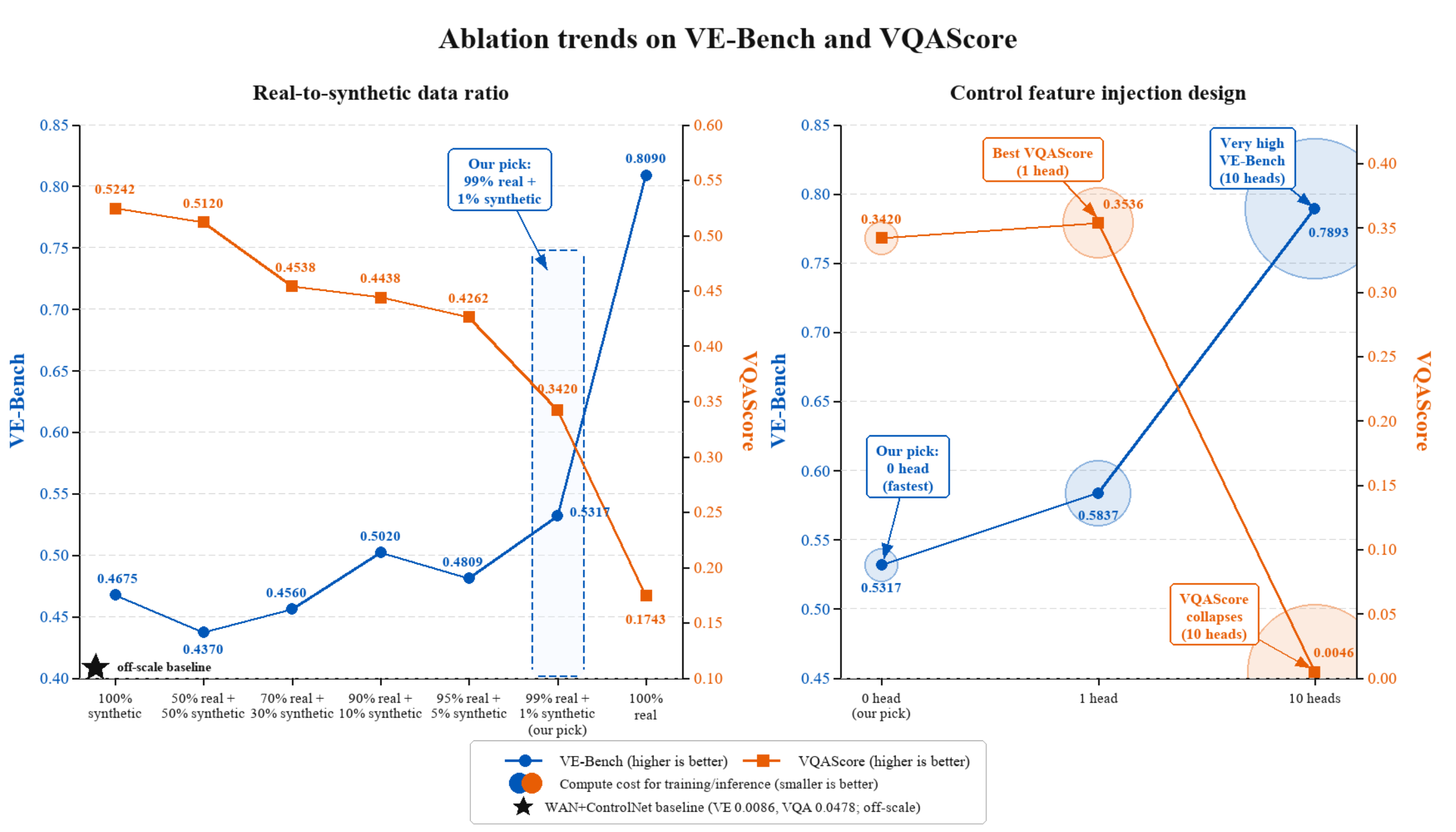}
  \end{minipage}

  \caption{\textbf{Quantitative ablation results.}
  We report VE-Bench and VQAScore for different real-to-synthetic sampling ratios and control feature injection designs.
  (a) Numerical results for both ablation studies.
  (b) Corresponding trend plot.
  Bold and underlined values indicate the best and second-best results, respectively.}
  \vspace{-0.5cm}
  \label{fig:mix_and_heads_combined}
\end{figure}


\subsubsection{Real-to-Synthetic Sampling Ratio}

We study how the real-to-synthetic sampling ratio in our domain-hedging strategy influences the trade-off between structural controllability (VE-Bench) and prompt adherence (VQAScore). For example, a ratio of 50\% real $+$ 50\% synthetic indicates that, for each training batch, samples are drawn from the real and synthetic data pools with equal probability. Figure~\ref{fig:mix_and_heads_combined} shows that increasing the proportion of real data generally improves controllability, while progressively reducing prompt adherence. In particular, strongly real-dominant mixtures such as 90\%/10\% and 99\%/1\% achieve substantially higher VE-Bench scores than synthetic-heavy settings, at the cost of lower VQAScores.

When training exclusively on real data (100\%), the model achieves the highest VE-Bench score but suffers a sharp drop in VQAScore, reflecting a loss of generative flexibility and a tendency to strictly reproduce the input rather than hallucinate novel content.

\begin{figure}[htbp]
    \centering
    \includegraphics[width=\linewidth]{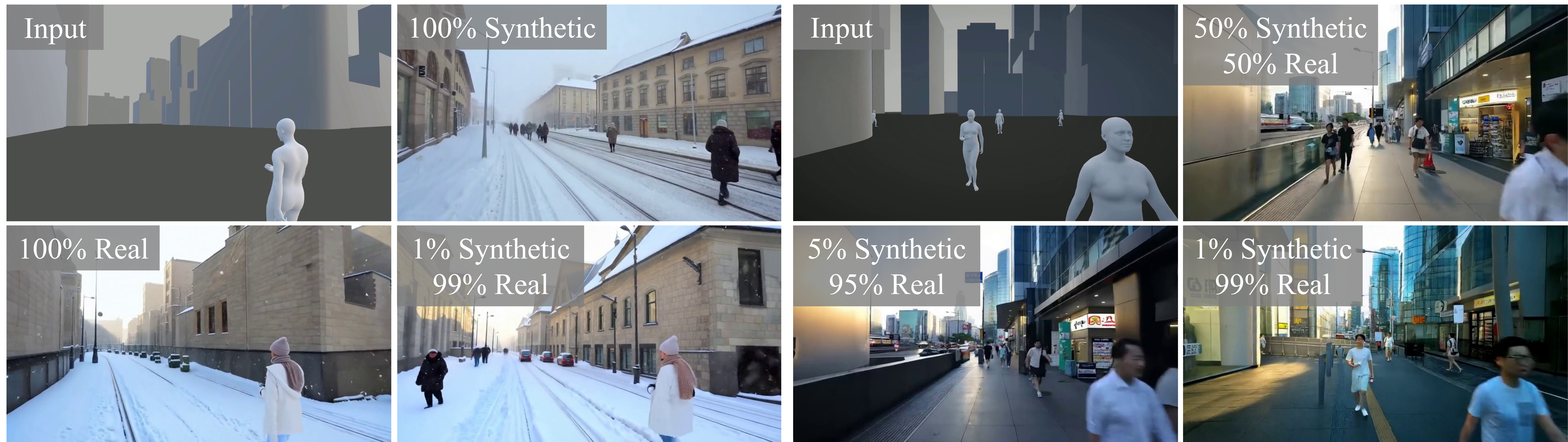}
    \caption{\textbf{Effect of the real-to-synthetic sampling ratio.} Synthetic-only training lacks diversity and weakly follows the control input, while real-only training improves structural alignment but reduces contextual richness. A strongly real-dominant ratio (99\% real / 1\% synthetic) provides the best trade-off, preserving control fidelity while enhancing detail generation. Increasing the proportion of synthetic data improves creativity but progressively weakens structural consistency, leading to hallucinations and deviations from the intended motion and layout.}
  \label{fig:data-ablation}
\end{figure}

This behavior highlights the central trade-off in our domain-hedging strategy: real data strengthens the photorealistic generative prior and improves structural fidelity, while synthetic data preserves controllability by anchoring the mapping between coarse inputs and target appearance. These trends are also evident in the qualitative results shown in Figure~\ref{fig:data-ablation}. Based on both quantitative and qualitative evaluations, we adopt a strongly real-dominant ratio of 99\% real and 1\% synthetic data.

\subsubsection{Control feature injection strategy.}

We inject DINO features from the control video into the first third of DiT blocks, which mainly influence structural control and global scene details. 
In Fig.~\ref{fig:heads} and Fig.~\ref{fig:mix_and_heads_combined}, we compare alternative but computationally heavier strategies for integrating control features via projection heads: (0) direct addition of DINO features to noisy latents, (1) a single shared projection control head, and (10) dedicated projection control heads (1 per block). Surprisingly, direct addition (0 heads) achieves comparable quality to the 1-head approach, demonstrating that no extra learned projection is needed, training only the Adapter projection of the DINOv3 features to the diffusion latent space is enough. Meanwhile, per-block heads (10 heads) provide excessive capacity, causing the model to memorize training inputs rather than generalize. This shows that simple feature addition is sufficient, and adding learnable capacity either provides no benefit or actively harms performance.

\begin{figure}[htbp]
  \centering
  \includegraphics[width=\linewidth]{images/pdfs/ablation_heads_v2.pdf}
\caption{\textbf{Ablation of control feature injection strategies.}
We compare direct additive injection after the DINO adapter (0 heads), a shared projection head (1 head), and per-block projection heads (10 heads). 
Direct addition is our preferred design, balancing control alignment and realistic detail synthesis while remaining computationally efficient, since it introduces no additional learnable parameters during training or inference. In contrast, additional heads increase conditioning capacity but can overfit the control signal and produce reconstruction-like outputs.}
  \label{fig:heads}
\end{figure}


\subsubsection{HSV Decorrelation}
\begin{figure}[htbp]
  \centering
  \includegraphics[width=\linewidth]{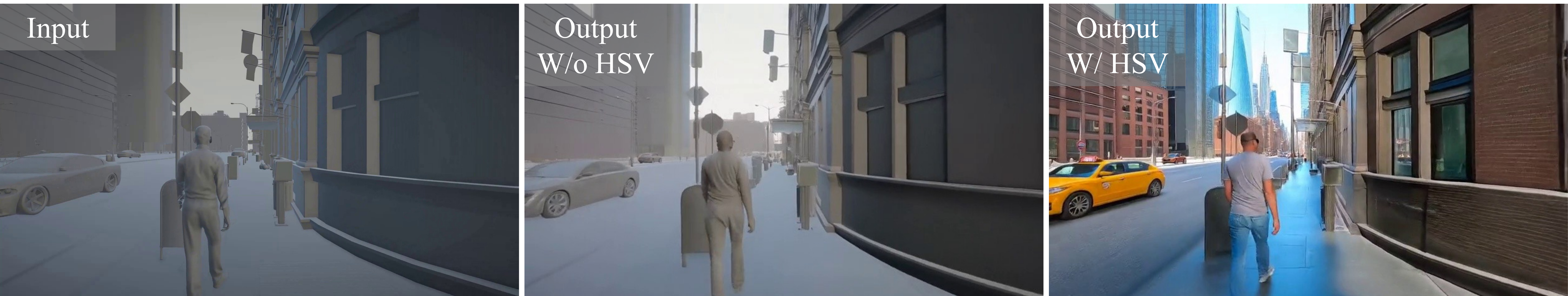}
  \caption{\textbf{Effect of HSV decorrelation in the control branch.} Without HSV decorrelation, the model inherits undesired appearance cues from the control input. Video-consistent HSV decorrelation suppresses color and texture leakage while preserving structure and temporal coherence.}
  \label{fig:nohsv}
\end{figure}

We study the effect of HSV decorrelation in the control branch during Stage II training. Without HSV decorrelation, control features retain low-level appearance cues, causing the model to inherit colors and textures from the input video, which is undesirable when appearance should be synthesized from text and learned priors. Applying a video-consistent HSV transformation suppresses appearance correlations while preserving structure and temporal coherence, encouraging the control signal to focus on spatio-temporal layout and preventing appearance leakage, as illustrated in Fig.~\ref{fig:nohsv}.

\subsection{Comparisons}
\label{sec:comparisons}

To evaluate the effectiveness of C2R, we compare our framework against several recent controllable video generation and video-to-video translation baselines, including WAN VACE~\cite{jiang2025vace,wan2025wanopenadvancedlargescale}, Control-A-Video~\cite{chen2023controlavideo}, Diffusion as Shader (DaS)~\cite{gu2025diffusionasshader}, WAN+ControlNet~\cite{wan2025wanopenadvancedlargescale,zhang2023adding,alibaba2025wanfun}, and Seedance 2.0~\cite{seedance2026seedance2}. These methods rely on different forms of conditioning, such as depth maps, tracking signals, or reference images, to guide video synthesis. In contrast, C2R operates directly on the coarse control video and text prompt, without requiring auxiliary geometric preprocessing or reference-image conditioning.

\begin{figure}[htbp]
  \centering
  \includegraphics[width=\linewidth]{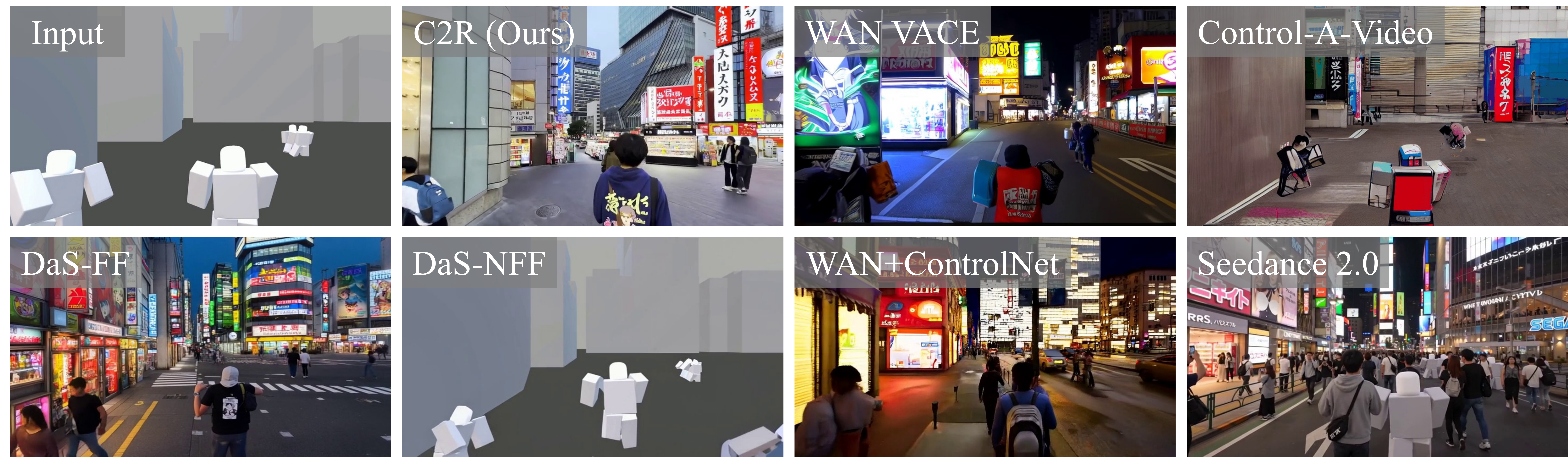}
  \caption{\textbf{Baseline comparison on coarse low-poly block-based character control.} We compare C2R against WAN VACE, Control-A-Video, Diffusion as Shader (DaS), WAN+ControlNet, and Seedance 2.0 under the same text prompt and coarse control input. Existing methods struggle to translate highly coarse low-poly block characters into realistic video outputs, often producing weakly controlled or reconstruction-like outputs. In contrast, C2R successfully transfers the coarse motion and layout into realistic humans and environments while preserving strong structural consistency and richer visual detail.}
  \label{fig:blockycomp}
  \vspace{-0.25cm}
\end{figure}

For a fair comparison, all methods are evaluated under the same target prompt and coarse low-poly block-based control setting shown in Fig.~\ref{fig:blockycomp}. WAN VACE uses depth sequences extracted from the input control video together with the text prompt, while Control-A-Video also relies on depth-conditioned video generation. Diffusion as Shader (DaS) requires additional tracking supervision computed with SpaTracker~\cite{xiao2024spatialtracker}, as well as a reference image. We evaluate two DaS variants: DaS-NoFirstFrame, which uses the first frame of the control sequence as reference, and DaS-FirstFrame, where the reference frame is generated with FLUX~\cite{blackforestlabs2024fluxdepth,blackforestlabs2024flux} following the original DaS pipeline. WAN+ControlNet corresponds to our finetuned WAN-ControlNet baseline trained on the same synthetic paired data as C2R, enabling direct conditioning on the input control video and text prompt without requiring depth or tracking extraction. Finally, we also compare against Seedance 2.0 using the same control video and prompt inputs.

As illustrated in Fig.~\ref{fig:blockycomp}, existing baselines struggle to translate highly coarse low-poly block characters into realistic video outputs. WAN VACE preserves the overall camera trajectory and scene structure, but its depth-based conditioning strongly constrains the generated humans to the coarse input geometry, resulting in rigid and insufficiently realistic appearances. Control-A-Video similarly preserves coarse identity cues, yet often fails to follow the intended motion and camera dynamics while producing low-quality generations. Reference-image-based approaches such as DaS improve visual fidelity when using a generated reference frame (DaS-FF), but still suffer from inaccurate motion alignment, while DaS-NFF often produces reconstruction-like outputs tied to the coarse input appearance. WAN+ControlNet partially enables coarse-to-real transfer while maintaining strong structural adherence, but remains visually conservative and lacks expressive detail synthesis. Seedance 2.0 generates highly realistic imagery, yet frequently fails to preserve the intended motion and scene alignment under coarse control inputs, occasionally introducing visible reconstruction artifacts from the block-based signal. In contrast, C2R successfully preserves layout, motion, and scene dynamics while generating photorealistic humans and environments with richer textures, lighting, and contextual detail.


To further explore these differences, Appendix Section~\ref{app:additional} presents the same evaluation protocol using humanoid control inputs instead of low-poly block characters. The results further show that C2R achieves a strong balance between visual richness, structural grounding, and motion alignment.

\section{Limitations and Future Work}
\label{sec:limitations}

Our method relies on coarse 3D inputs to guide scene layout, camera motion, and human dynamics. When the input geometry is extremely sparse or abstract, it may not provide sufficient structural cues to precisely control the desired layout or camera trajectory. In such cases, text prompts alone may be ambiguous and insufficient to fully disambiguate camera motion or character movement, as illustrated in the Appendix Section~\ref{app:limitations}. Future work includes incorporating more explicit control signals, such as camera motion directions, speed profiles, or character movement constraints, to improve controllability. Additionally, the current framework operates in a non-autoregressive manner; extending the model to an autoregressive formulation could enable real-time generation and interactive applications.

\section{Acknowledgement}
We thank Super Dimension (Jiaxing) InfoTech Ltd. for providing the 3D scanned human models used in this work. We also thank Dave Cardwell for the initial and insightful discussions on production-level requirements for crowd simulation; this project originated from those early conversations. This work was partially funded by the project 'Inteligencia artificial para la industria 4.0: generación de datos, modelado avanzado optimización e interpretabilidad' (IDEA-CM), with reference TEC-2024/COM-89, funded by the Community of Madrid through the call for grants for collaboration R\&D projects in the 2024 Technology R\&D Activity Programs category, according to Order 3177/2024.


\bibliographystyle{plainnat}
\bibliography{references/references}


\clearpage
\newpage

\appendix
\section{Appendix}
\label{sec:appendix}

\subsection{Implementation Details}

\subsubsection{Training Protocol}

We follow the standard training protocol of latent diffusion and flow-matching video generation models. Videos are temporally sampled to 81 frames at 16 fps, resized and center-cropped to the $832 \times 480$ training resolution, and encoded into the latent space of the pretrained Wan 2.1 VAE. Text prompts are encoded with the frozen T5-XXL text encoder. At each training step, we sample a timestep $t$ from the flow-matching schedule, perturb the clean latent $\mathbf{z}_0$ with Gaussian noise to obtain $\mathbf{z}_t$, and optimize the model to predict the corresponding velocity field using the objective in Eq.~\ref{eq:diff_objective}.

Training is performed in two stages. In Stage I, we fine-tune the diffusion backbone on real videos to adapt the pretrained model to our target urban video distribution, while keeping the VAE and text encoder frozen. In Stage II, we freeze the Stage-I backbone and train only the control pathway, namely the DINO feature adapter that projects temporally concatenated frame-wise features into the diffusion latent space. The DINO encoder itself remains frozen. Stage II uses the synthetic-real domain-hedging strategy described in Section~\ref{subsec:domain_hedging}, where paired synthetic data provides explicit coarse-to-real supervision and large-scale real videos regularize the control pathway toward the photorealistic distribution learned in Stage I.

We split the data into training and held-out validation sets using a 90--10 ratio for both the real and the synthetic data, ensuring that videos from the same source sequence do not appear in both splits. The validation set is used to monitor convergence, compare ablations, and inspect fixed validation generations throughout training, while final results are computed on a separate evaluation set. Although the flow-matching validation loss provides a useful convergence signal, checkpoint selection is based on both validation-loss trends and qualitative assessment of the realism--controllability trade-off, including temporal consistency, prompt alignment, control adherence, and appearance leakage.

We use an effective global batch size of 32. Both stages are optimized with AdamW using mixed-precision distributed training. Stage I uses a learning rate of $1 \times 10^{-5}$, while Stage II uses a learning rate of $1 \times 10^{-4}$. All ablations use the same optimizer settings, training resolution, data split, and validation protocol for fair comparison. Checkpoints are saved periodically during training. The complete two-stage training process takes approximately three weeks on a single node with eight NVIDIA H100 GPUs using data-parallel training.

\subsubsection{Shared Feature Space}

\begin{figure}[htpb]
  \centering
  \includegraphics[width=\linewidth]{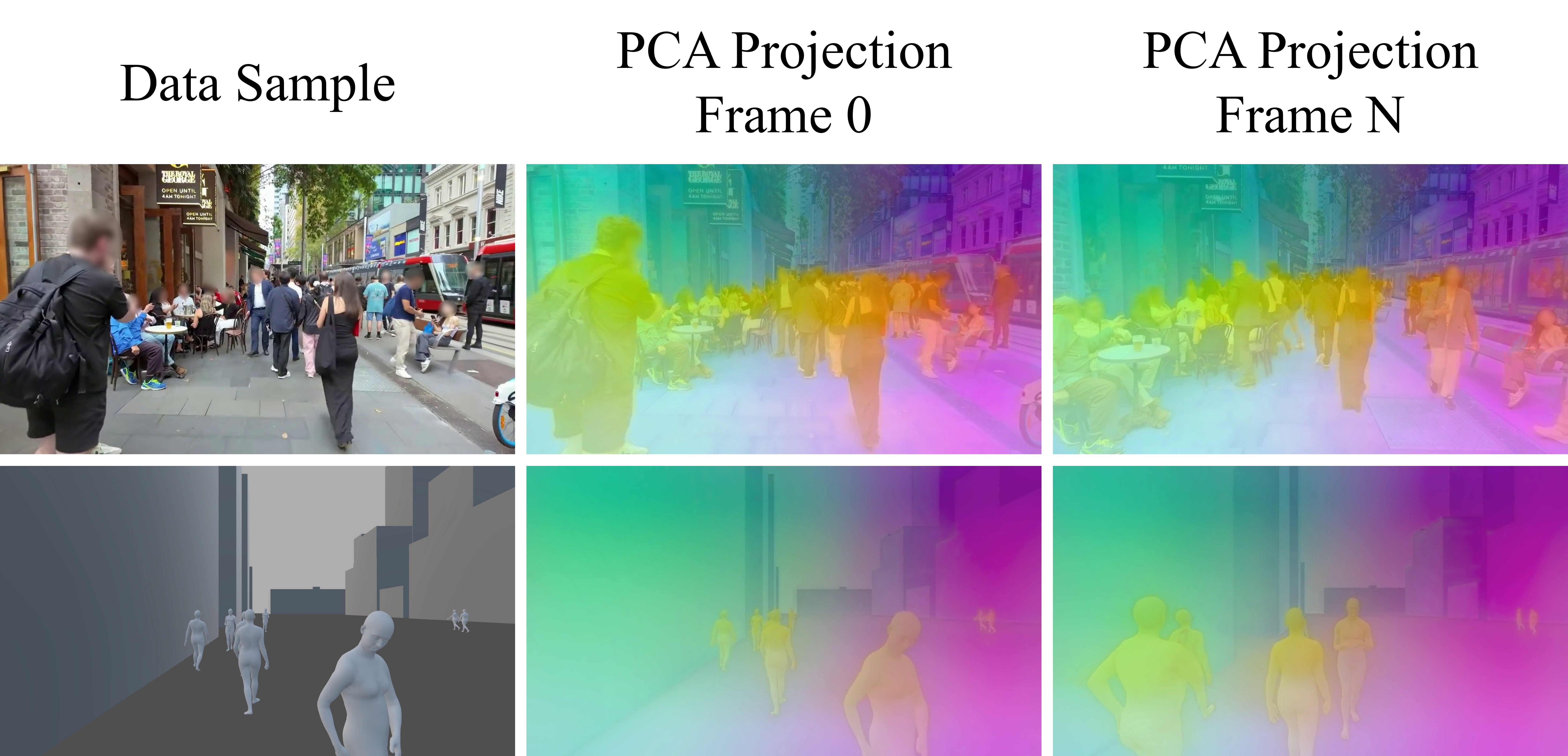}
  \caption{\textbf{DINOv3 features provide a domain-robust and temporally stable control signal.} We visualize patch embeddings using a \emph{global} PCA projection computed on a mixed subset of real and synthetic samples, then reused across videos. Similar PCA colors across real and synthetic inputs indicate that DINOv3 aligns corresponding structural elements despite large appearance gaps. Color stability across time suggests temporal coherence even when features are extracted per-frame, supporting spatio-temporal control.}
  \vspace{-0.5cm}
  \label{fig:dino_viz}
\end{figure}

In Fig.~\ref{fig:dino_viz}, we demonstrate how DINOv3 features serve as the foundation for a common spatio-temporal feature space that bridges the domain gap between unpaired real videos and paired synthetic data. By mapping both real videos and coarse 3D renders into this shared representation, our pipeline ensures that control signals from vastly different origins are structurally compatible for joint training. The visualization of patch embeddings via a global PCA projection reveals that DINOv3 aligns corresponding semantic elements across the real and synthetic domains, effectively neutralizing appearance gaps. This shared space allows paired synthetic data to act as a structural anchor linking geometry to target appearance, while real videos populate the space to enrich the learned generative distribution.

\subsubsection{Preventing Feature Leakage with Video-Consistent HSV Decorrelation}
\label{paragraph:hsv_decorrelation}

\textbf{Problem.} DINO features are expressive: if the control branch receives the same pixels as the VAE branch, the adapter can pass appearance information that encourages reconstruction rather than synthesis.

\textbf{Solution.} For real videos during Stage II, we feed the same video $\mathbf{x}$ to both branches, but apply a \emph{video-consistent} random HSV transformation only to the control branch:
\begin{equation}
\mathbf{x}_{\text{ctrl}} = \mathrm{HSV}(\mathbf{x}),
\end{equation}
and compute $\hat{\mathbf{z}}_{\text{ctrl}} = A(\mathrm{DINO}(\mathbf{x}_{\text{ctrl}}))$.
By randomly shifting hue and scaling saturation/value consistently across frames, the adapter is forced to prioritize geometry and dynamics over pixel-level appearance. Compared to grayscale, HSV decorrelation avoids systematically biasing outputs toward muted colors and better preserves realistic color diversity. No augmentation is required at inference time.

In a nutshell, we apply a temporally consistent HSV decorrelation augmentation independently to each training video. For every video, a single random HSV transformation is sampled and applied uniformly across all frames to preserve temporal coherence while altering appearance statistics between videos. Specifically, we randomly sample a hue offset with magnitude between \(20^\circ\) and \(90^\circ\), with equal probability of positive or negative rotation. In addition, saturation is randomly scaled by \(\pm 15\%\) to \(\pm 40\%\), and value (brightness) by \(\pm 15\%\) to \(\pm 30\%\). Each frame is converted from RGB to HSV space, transformed using the same per-video hue offset and saturation/value scaling factors, clipped to the valid range, and converted back to RGB before being used for training. This augmentation reduces appearance leakage while preserving the underlying structure and motion consistency of the control signal.

\subsubsection{Post-Processing and Artifact Mitigation}

While our foundational Wan 14B model occasionally generates frames containing visible watermarks, we address this through an automated post-processing pipeline. Specifically, we employ specialized content-aware restoration tools to systematically detect and remove these artifacts, ensuring that the final synthesized videos maintain a clean and professional visual quality.

\subsection{Existing Video Models for Populated Urban Scene Generation}
\label{app:existing_models}

\begin{figure}[htbp]
  \centering
  \includegraphics[width=\linewidth]{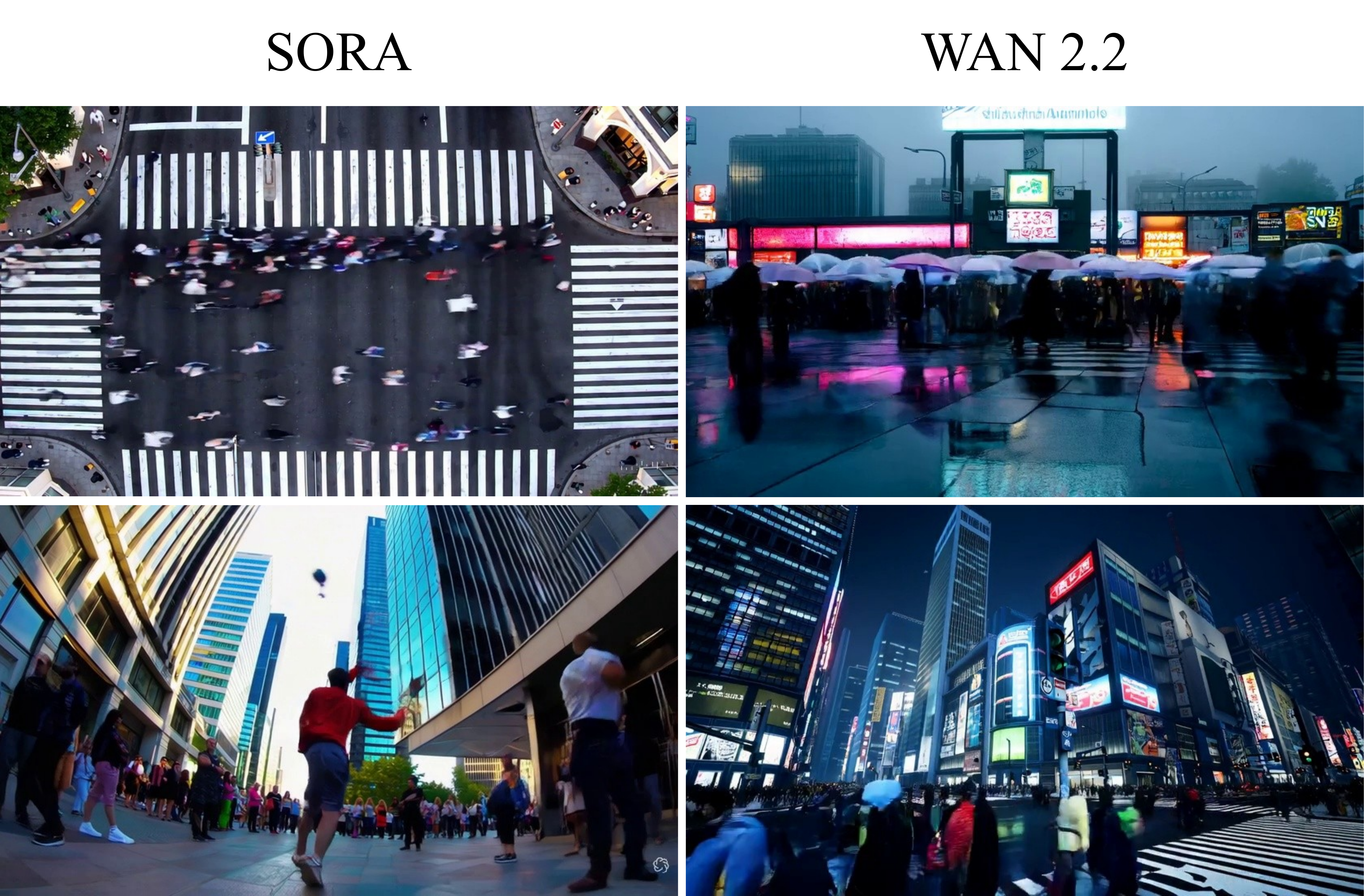}
  \caption{SORA (left column) and WAN (right column) show limited controllability over human motion and camera trajectories, and tend to generate similar viewing angles across populated urban scenes.
  }
  \vspace{-0.5cm}
  \label{fig:defaults}
\end{figure}

Fig.~\ref{fig:defaults} illustrates the difficulty of generating scenes similar to those presented in our experiments using traditional text-to-video models without any control signal, even when provided with detailed prompts. In particular, prompts involving cities, crowds, or general human activities often lead to fast-forward-like scene dynamics, unstable motion, and limited temporal consistency. Moreover, character identity, layout, and camera trajectories remain extremely difficult to control reliably in purely prompt-driven generation. These limitations motivate the need for controllable video generation frameworks capable of leveraging the strong generative priors and world knowledge of large-scale video models while grounding them toward specific tasks, scenarios, and structural constraints.

\subsection{Data Collection}
\label{app:data}


We curate a large-scale dataset consisting of both synthetic and real-world videos to support training and evaluation.

For synthetic data generation, we use professional 3D content creation tools to simulate humans interacting within complex urban 3D environments. The scenes are of AAA visual quality, composed of fully textured city models with realistic geometry, materials, lighting, and post-processing effects, and populated with high-fidelity human assets obtained from detailed 3D scans \cite{superdimension}. City environments were obtained from Sketchfab and created by Abhayexe under a Creative Commons license \cite{abhayexe_sketchfab}. To enable paired supervision, each city environment is additionally converted into a corresponding coarse representation by approximating the volume of every building with simple flat geometric primitives (e.g., cuboids), producing a simplified city layout that preserves large-scale structure while removing fine visual detail. Vehicles are treated similarly, being replaced in the coarse domain by simple, untextured meshes that approximate their overall volume without appearance cues.

For character pairing, each high-quality scanned human model is matched with a coarse counterpart consisting of a simple unclothed human body mesh without hair or accessories. 
Both the full-quality and coarse characters are driven by the same motion data by retargeting identical Mixamo animations \cite{mixamo}, ensuring precise alignment of motion, pose dynamics, and overall body structure across representations. This pairing strategy guarantees that differences between the full and coarse renders arise solely from appearance and geometric detail, rather than from motion discrepancies.

For each sample in the synthetic dataset, we randomly spawn a crowd of characters in a street region of the city, assign randomized textures and animations to the full-quality assets, and propagate the corresponding animations to their coarse counterparts. Two cameras with identical trajectories are then attached to a randomly selected character in the crowd and used to record a short replay sequence of 5 seconds. The full camera exclusively captures the high-quality assets, including detailed city geometry, vehicles, scanned characters, lighting, and post-processing effects, while the coarse camera records only the simplified elements: unclothed character body meshes, a ground plane, and the coarse volumetric city and vehicle representations rendered with a uniform, neutral material (e.g., grayscale). Each iteration produces one paired data sample consisting of a full-quality video and its corresponding coarse video, enabling learning of fine-grained visual detail from structured but minimal 3D representations.

For real data, we collect street-view video footage from cities spanning all five continents, covering a broad variety of urban environments. The data includes a wide range of viewpoints and motion characteristics, such as static and dynamic captures, first- and third-person perspectives, and varying viewing directions. We segment the videos into short clips to facilitate training.


We automatically generate textual annotations for both real-world videos and full-quality synthetic clips using Tarsier \cite{yuan2025tarsier2advancinglargevisionlanguage}, a state-of-the-art video captioning system built on top of the Qwen large multimodal model~\cite{bai2023qwen}. For each video, we employ a simple and generic prompt (e.g., “Describe the video in detail”), relying on the strong video–language understanding capabilities of Tarsier to produce rich, free-form captions without manual engineering of task-specific prompts.

The resulting captions consistently capture high-level scene context and fine-grained semantic attributes, including human actions and motion patterns, camera motion, clothing appearance, crowd density, weather conditions, environmental layout, architectural style, and location cues derived from the appearance of the city. This process yields expressive and diverse textual descriptions that reflect realistic real-world semantics and visual variability.

We apply this automatic captioning procedure to all real video clips in the dataset as well as to the full-quality synthetic renders. Coarse synthetic videos are not captioned separately; instead, each coarse clip inherits the caption of its corresponding full-quality counterpart. Since both representations are perfectly paired in terms of motion, structure, and scene layout, the caption associated with the realistic render provides an accurate semantic description for the paired coarse input. This design allows the model to learn to condition generation on realistic, high-level textual supervision while operating on minimal and abstract 3D visual representations.

In total, our dataset contains 240K real-world video clips and 1.3K synthetic video clips, each with 5 seconds sampled at 16 fps.

\subsection{Extra Experiments}
\subsubsection{Additional Comparison Examples}
\label{app:additional}

\begin{figure}[htpb]
  \centering
  \includegraphics[width=\linewidth]{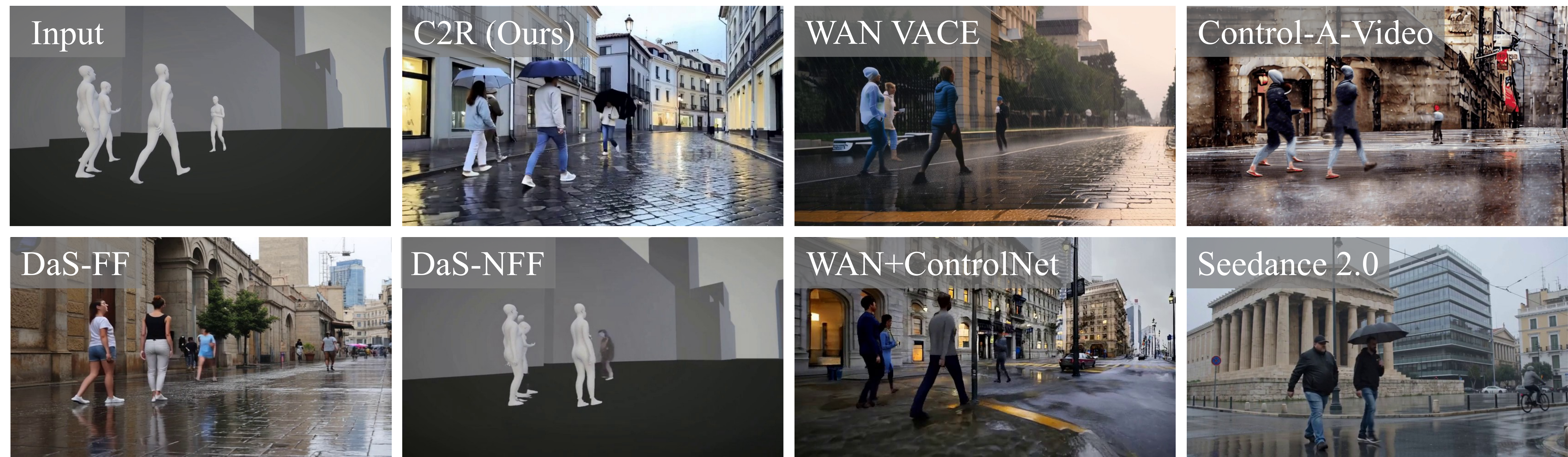}
  \caption{\textbf{Baseline comparison on coarse humanoid character control.} We compare C2R against WAN VACE, Control-A-Video, Diffusion as Shader (DaS), WAN+ControlNet, and Seedance 2.0 under the same target prompt and humanoid control setting. While the humanoid input provides stronger structural priors than the low-poly block scenario, existing baselines still struggle to accurately preserve motion structure, camera dynamics, and temporal coherence across the generated sequence. In contrast, C2R achieves the best balance between structural controllability and photorealistic synthesis, generating realistic humans and environments while faithfully following the input motion and layout.}
  \label{fig:humanoidcomp}
\end{figure}

To further evaluate the performance of C2R, we provide an extended qualitative comparison in Figure~\ref{fig:humanoidcomp} using coarse humanoid control inputs instead of low-poly block characters. Following the same evaluation methodology as in Section~\ref{sec:comparisons}, we compare C2R against several recent controllable video generation baselines, including WAN VACE, Control-A-Video, Diffusion as Shader (DaS), WAN+ControlNet, and Seedance 2.0, under the same target prompt and control setting.

Similarly to the block-based comparison, the evaluated methods rely on different conditioning modalities to guide generation. WAN VACE and Control-A-Video use depth-conditioned video generation, while DaS additionally requires tracking supervision computed with SpaTracker together with a reference image. We evaluate both DaS-NoFirstFrame, which uses the first frame of the control sequence as reference, and DaS-FirstFrame, where the reference frame is generated using FLUX following the original DaS pipeline. WAN+ControlNet corresponds to our finetuned WAN-ControlNet baseline trained on the same synthetic paired data as C2R, enabling direct conditioning on the humanoid control sequence and text prompt. Seedance 2.0 is also evaluated using the same control video and prompt inputs.

As illustrated in Fig.~\ref{fig:humanoidcomp}, the humanoid control setting is significantly easier than the highly coarse low-poly block scenario, since the input already provides more human-like structural priors and motion cues. Nevertheless, important differences between methods remain apparent. Depth- and tracking-based approaches generally maintain the input motion structure and coarse character consistency, although they can still fail to preserve fine identity details across the sequence. Their outputs often remain visually conservative, with limited photorealism and weak environmental dynamics, where backgrounds remain largely static despite foreground motion. While reference-image-based methods improve visual fidelity, they still inherit many of the limitations of depth- and tracking-conditioned generation, resulting in constrained scene evolution and less expressive video synthesis. WAN+ControlNet improves overall structural and temporal adherence compared to off-the-shelf controllable baselines, but still produces visually conservative outputs with limited realism and detail synthesis. Seedance 2.0 generates visually appealing frames, yet frequently fails to accurately follow the intended camera trajectory and motion structure.

In contrast, C2R successfully preserves the humanoid motion, camera dynamics, and scene layout while generating realistic humans, detailed clothing, rich urban textures, and coherent lighting conditions, even hallucinating fine contextual details such as umbrellas that are not explicitly provided in the coarse control input. Compared to the low-poly block experiments, all methods benefit from the more informative humanoid control input; however, C2R consistently achieves the best balance between structural controllability and photorealistic synthesis. These results further support the effectiveness of our synthetic-real domain-hedging strategy across varying levels of control signal abstraction.


\subsubsection{Adaptive Coarse-to-Fine control}
\begin{figure}[htbp]
  \centering
  \includegraphics[width=0.85\linewidth]{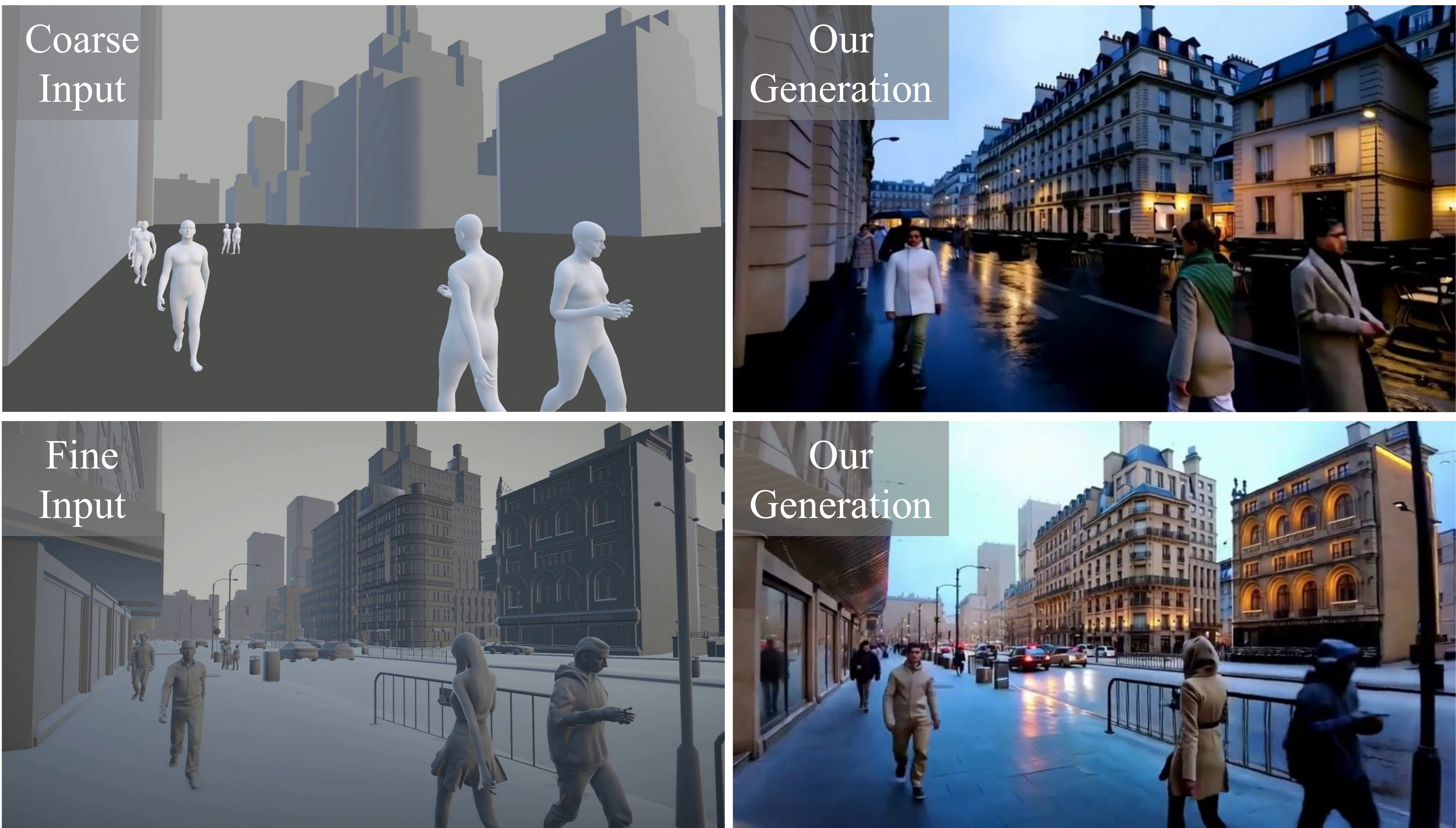}
  \caption{\textbf{Coarsening levels for driving signal.} 
  Given very coarse geometry (top), our model inpaints many details. Given fine geometry (bottom), it successfully follows the richer input signal.
}
  \label{fig:coarse2fine}
\end{figure}

Fig.~\ref{fig:coarse2fine} shows that our method adapts control signals at different levels of geometric detail, from very coarse to relatively fine inputs. In all cases, the model respects the provided structure and adapts the generated content to the fidelity of the input geometry.

\subsubsection{Applications}
Our framework is agnostic to the specific human and scene templates used in the control video and generalizes well beyond the training distribution. By operating on coarse 3D renderings and applying HSV augmentation during training, the model reduces dependence on appearance cues such as color, texture, and rendering style. This allows robust handling of inputs from different game engines, simulation pipelines, and lighting conditions while preserving camera motion and human trajectories.

As shown in Fig.~\ref{fig:application_roblox}, our method transforms low-poly game videos into realistic outputs while faithfully following the input motion and interactions. Notably, the input character is cartoon-styled and visually far from the neutral human models used during training, yet the generated video reproduces the same running and jumping motion with improved realism. Visual collision artifacts present in the low-poly input are also mitigated, resulting in more plausible contact between the character and the environment.

\begin{figure}[htpb]
  \centering
  \includegraphics[width=\linewidth]{images/pdfs/app_roblox_1_v2.pdf}
  \caption{\textbf{Turn a low-poly Roblox game video into real-style.} \emph{Left:} The character runs forward and jumps over a low wall. \emph{Right:} The character throws a bomb and runs away.}
  \label{fig:application_roblox}
\end{figure}

\subsubsection{Illustrating Limitations}
\label{app:limitations}

\begin{figure}[htbp]
  \centering
  \includegraphics[width=\linewidth]{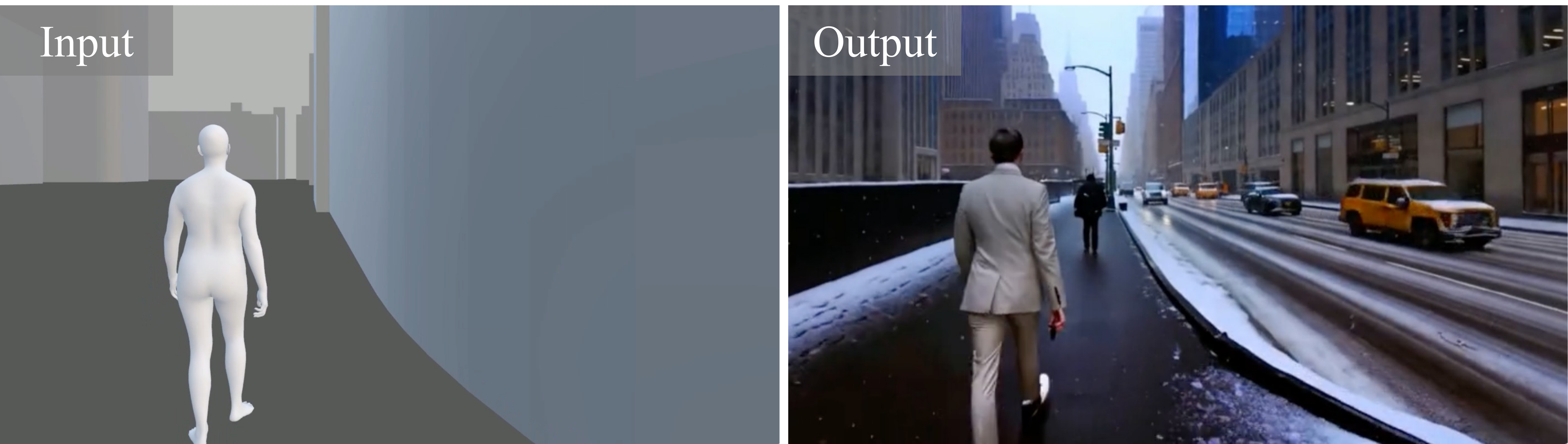}
  \caption{\textbf{Limitation.} Very coarse input 3D structure might not be sufficient to guide the architecture and city layout as expected.}
  \label{fig:limitation}
\end{figure}

Fig.~\ref{fig:limitation} illustrates how our method can fail to maintain structural integrity when relying on extremely sparse or abstract 3D inputs. While these coarse inputs are intended to guide scene layout, camera motion, and human dynamics, a lack of sufficient geometric detail can result in a loss of precise control over the desired layout or trajectory.

\end{document}